# Fossil image identification using deep learning ensembles of data augmented multiviews

Chengbin Hou[1,2] | Xinyu Lin[2,3] | Hanhui Huang[4] | Sheng Xu[3] | Junxuan Fan[4] | Yukun Shi[4] | Hairong Lv[1,2]

[1]Ministry of Education Key Laboratory of Bioinformatics, Bioinformatics Division, Beijing National Research Center for Information Science and Technology, Department of Automation, Tsinghua University, Beijing, China

[2]Fuzhou Institute of Data Technology, Fuzhou, China

[3]College of Physics and Information Engineering, Fuzhou University, Fuzhou, China

[4]School of Earth Sciences and Engineering and Frontiers Science Center for Critical Earth Material Cycling, Nanjing University, Nanjing, China

**Correspondence**
Yukun Shi
Email: ykshi@nju.edu.cn

Hairong Lv
Email: lvhairong@tsinghua.edu.cn

**Funding information**
National Natural Science Foundation of China, Grant/Award Number: 42050101 and 42250104; National Key R&D Program of China, Grant/Award Number: 2021YFB3600401; Fujian Provincial Natural Science Foundation, Grant/Award Number: 2021J01586; Deep-time Digital Earth (DDE) Big Science Program.

**Handling Editor:** Arthur Porto

## Abstract

1. Identification of fossil species is crucial to evolutionary studies. Recent advances from deep learning have shown promising prospects in fossil image identification. However, the quantity and quality of labelled fossil images are often limited due to fossil preservation, conditioned sampling and expensive and inconsistent label annotation by domain experts, which pose great challenges to training deep learning-based image classification models.

2. To address these challenges, we follow the idea of the wisdom of crowds and propose a multiview ensemble framework, which collects Original (O), Grey (G) and Skeleton (S) views of each fossil image reflecting its different characteristics to train multiple base models, and then makes the final decision via soft voting.

3. Experiments on the largest fusulinid dataset with 2400 images show that the proposed OGS consistently outperforms baselines (using a single model for each view), and obtains superior or comparable performance compared to OOO (using three base models for three the same Original views). Besides, as the training data decreases, the proposed framework achieves more gains. While considering the identification consistency estimation with respect to human experts, OGS receives the highest agreement with the original labels of dataset and with the re-identifications of two human experts. The validation performance provides a quantitative estimation of consistency across different experts and genera.

4. We conclude that the proposed framework can present state-of-the-art performance in the fusulinid fossil identification case study. This framework is designed for general fossil identification and it is expected to see applications to other fossil datasets in future work. Notably, the result, which shows more performance gains as train set size decreases or over a smaller imbalance fossil dataset, suggests the potential application to identify rare fossil images. The proposed framework also demonstrates its potential for assessing and resolving inconsistencies in fossil identification.

Chengbin Hou, Xinyu Lin and Hanhui Huang contributed equally to this study.

**KEYWORDS**
deep learning, ensemble, fossil identification, fusulinid, identification inconsistency, image classification, palaeoecology

## 1 | INTRODUCTION

Evolutionary studies require the accurate and efficient identification of extant and especially fossil species, however, the hope is often frustrated by several restrictions. Most fossil species and many extant species are defined by their phenetic characters, and thus type specimens need to be assigned to represent their typical morphology. Due to the limited accessibility of type specimens, identification often relies on images, which presents challenges for researchers. This problem is particularly serious for palaeontologists, as fossil species assignment is usually based on a small number of samples in varying states of preservation (Behrensmeyer et al., 2000; Foote & Raup, 1996; Holland, 2016; Schopf, 1975), compared to extant species samples that are more abundant and readily available. Moreover, as research in the life and earth sciences tends to assemble more data for larger-scale, higher-resolution studies, the relatively small community of taxonomists has to spend a large amount of time and effort in routine identification tasks, and is thus hindering broader taxonomic studies (MacLeod et al., 2007, 2010). These issues highlight the increasing need for auxiliary tools or automatic identification systems to aid taxonomists in improving the efficiency and accuracy of their identification and making large-scale studies with well-identified samples feasible.

Automatic identification models have remained heavily practiced in current biological and ecological studies for years, with numerous studies focusing on the identification of extant species (Borowiec et al., 2022; Wäldchen & Mäder, 2018); however, there has been less focus on applying them to the studies of deep time. Fossil species are as rich in morphological diversity as modern organisms, but the available material is severely limited by fossil preservation and sampling intensity, which would affect the model training. With a limited number of samples, the model may not be able to fully learn the differences in features across categories, making it challenging to train effectively, or it may overfit, resulting in poor performance on the test set or in real-world applications. Another concern is the quality of fossil images, which is typically worse than that of modern species because the formation, burial and sampling conditions of fossils can greatly alter the images, posing a greater challenge to the recognition ability of the model. The problem also lies in the labelling process. Taxonomic and systematic studies of some fossil groups, mostly relying on limited morphological information due to the general lack of molecular data, are insufficient and sometimes contradictory. This could lead to disagreement among experts, causing inconsistency in data annotation and affecting the training of supervised learning models.

Nonetheless, recent advances in the use of deep learning models for taxonomic identification have shown promising prospects for the application on fossil taxa, including foraminifera (Hsiang et al., 2019; Marchant et al., 2020; Mitra et al., 2019; Pires de Lima et al., 2020), graptolites (Niu & Xu, 2022), fossil leaves (Wilf et al., 2021), pollen (Punyasena et al., 2022) and multiple-body-fossil mixture (Liu et al., 2022). The identification of modern foraminifera could be well compared to that of fossil foraminifera due to their close morphology and modes of preservation, and they are also among the first to be tested for species identification using deep learning. In their excellent study, Hsiang et al. (2019) constructed a large image dataset of over 34,000 planktonic foraminifera and used most of these images to train three commonly used neural networks, VGG-16, DenseNet-121 and Inception-v3. The species-level identification achieved a maximum accuracy of 87.4%, which is comparable to the expert accuracy of 63%–85%. In another study also conducted on modern foraminifera, Mitra et al. (2019) performed a more systematic comparison of human expert versus machine performance. Their results reveal that the combination of VGG-16 and ResNet-50 neural networks could achieve an accuracy of at least 80%, while the performance of 11 human identifiers varied dramatically with an average accuracy of 63%. These studies show the promise of deep learning for fossil species identification. However, many thorny issues might arise as the categories and ages of fossils expand.

To delve into the automatic identification of fossils, we take fusulinids, a large group of fossil foraminifera dating back to c. 300 Ma, as the subject of our study. Fusulinids are the earliest larger benthic foraminifera that appeared in the shallow water of the Carboniferous and survived until the Late Permian (Pawlowski et al., 2003; Vachard et al., 2010). Their rapid evolution, as seen in morphological changes, makes them prominent index fossils for the Late Palaeozoic biostratigraphy, that is, dating the bearing rocks (BouDagher-Fadel, 2008; Ross & Ross, 1991). Unlike modern foraminifera, fusulinids are primarily preserved in rocks that are difficult to separate, and studies are typically conducted on thin slices of the fossils that have been professionally made from rocks. This procedure compresses the three-dimensional morphological features into two dimensions, which is also common in the studies of other fossils such as corals, brachiopods, archaeocyathids, plants and even vertebrate bones. To meet different research needs, multiple sections of fusulinid fossils, including axial, sagittal and tangential sections, could be produced, and axial sections are preferred for identification as they contain the most useful features (Sheng et al., 1988; Vachard et al., 2010). This slice-based identification of fusulinids is very beneficial for applying automatic identification models, as deep learning models based on two-dimensional images have already been well developed. The use of deep learning on fusulinid identification is a rather unexplored subject, and, to our knowledge, only one study by Pires de Lima et al. (2020) serves as an example. They collected images of fusulinids from thin-slice micrographs and literature to construct a dataset containing 342 images of eight genera. Five standard neural network models (VGG-19, Inception-v3, MobileNet-v2, ResNet-50, and DenseNet-121) were trained on their dataset using transfer learning, and a maximum accuracy of 89% was achieved on Inception-v3 (Pires de Lima et al., 2020). Although the dataset they used is small and has a rather uneven distribution of categories (the smallest category has only 15 images, while the largest has 88), it still provides extremely valuable feasibility validation. To investigate the effectiveness of the newly proposed method in this study, we utilize the largest dataset of fusulinids to date (Huang et al., 2023), containing 2400 images from 16 genera that cover all six fusulinid subfamilies with respect to the classification system of Sheng et al. (1988).

Distinguished from the previous fossil identification studies that directly apply the existing machine learning and deep learning models, we follow the idea of the wisdom of crowds and propose a multiview ensemble framework (i.e. a kind of meta-method) to further improve the performance of existing deep learning models. Specifically, to compensate for the image quality and sample size, the fossil images in the original form are transformed into other fossil identification preferred forms to highlight various features of the same fossil image from diverse views. The diverse views of training images with their labels are then fed respectively to train multiple base models, and the predictions from these base models are combined to provide the final predictions. According to the proposed framework and the characteristics of fusulinids, we develop the OGS method that feeds the Original (O), Grey (G) and Skeleton (S) views of fusulinid images to three base models respectively. We select several milestone models in deep learning (ResNet, MobileNet, Inception, EfficientNet and RegNet) as the base model to validate the effectiveness of the proposed framework and the OGS method.

The main novelty of the proposed framework lies in the input to each base model. The choice of diverse input views depends on the characteristics of concrete applications, for example, the Grey view may help filter colour noise, and the Skeleton view could help extract morphological features, as demonstrated in this study to identify fusulinid fossil images. On the one hand, the proposed framework is motivated by the bagging framework (Dong et al., 2020; Zhou, 2021) in the field of ensemble learning, which is rarely used in fossil image classification. The main difference is that the bagging framework takes random samples from original images as the input to each base model. On the other hand, the proposed method is also inspired by the recent advances using data augmentation and ensemble to boost performance (He et al., 2016; Shorten & Khoshgoftaar, 2019; Simonyan & Zisserman, 2015). However, these deep learning studies often perform several data augmentation techniques (to enrich samples) to train or infer over a single model respectively, and an ensemble technique is employed to combine the outputs from that single model during testing. Note that the proposed method trains and infers over multiple base models using fossil-dedicated and meaningful views rather than commonly used data augmentation techniques such as resize, crop and flip.

The main contributions of this work as follows. First, we propose a multiview ensemble framework rather than a specific method, which can be broadly applied to various deep learning image classification models. Second, considering the characteristics of fossil images, we suggest the Grey and Skeleton views for data augmentation and accordingly develop the OGS ensemble. Extensive experiments on two fossil image datasets over five representative deep learning models are conducted to demonstrate the merits of OGS method. Third, further consistency experiments involving OGS models and human experts are performed, and the inconsistency among human-given labels is analysed. The results show the potential of using OGS model to assess and resolve identification inconsistency. Finally, the source code (Hou et al., 2023) is publicly available at https://github.com/houchengbin/Fossil-Image-Identification to benefit future research in fossil image identification.

## 2 | MATERIALS AND METHODS

### 2.1 | Dataset

The main dataset used in this work is 'Fusulinid images 2400 - NJU', which is described in detail in Huang et al. (2023) and hereafter referred to as the Huang et al. (2023) dataset. It is available for download at DDE repository at https://doi.org/10.12297/dpr.dde.202211.5. It consists of 2400 thin-slice images of fusulinid individuals, including 295 microscope photos and 2105 scanned images from the literature. The images are stored as PNG files with the transparency channel annotating the outline of the fossils and labelled according to their species name and data source. The 2400 images are selected evenly from 16 genera of all six fusulinid families: Fusulinidae, Schwagerinidae, Ozawainellidae, Schubertellidae, Neoschwagerinidae and Verbeekinidae (see Table 1). Images of holotypes, paratypes, cotypes and syntypes of the selected species are preferably chosen as they better represent the described morphological features. Although the images are labelled to the species level, using this level would result in a significantly imbalanced data volume, so the genus level was chosen in our study. In the main experiments, the Huang et al. (2023) dataset is split into a training set, a validation set and a test set, with 110, 20 and 20 images for each genus as the default setting. Other split ratios over this dataset are also carefully examined, as shown in Section 3.2 and Figure 2.

### 2.2 | Problem formulation

The fossil identification problem is a typical multiclass image classification problem. Considering a fossil dataset with totally $k$ categories, the aim is to build a classification model $f$ such

**TABLE 1** Overview of the taxonomy and number of images in the Huang et al. (2023) dataset.

| Family | Subfamily | Genus | Number of images |
| --- | --- | --- | --- |
| Fusulinidae | Fusulininae | *Fusulina* | 150 |
| Fusulinidae | Fusulinellinae | *Fusulinella* | 150 |
| Fusulinidae | Staffellinae | *Nankinella* | 150 |
| Schwagerinidae | Schwagerininae | *Chusenella* | 150 |
| Schwagerinidae | Schwagerininae | *Eoparafusulina* | 150 |
| Schwagerinidae | Schwagerininae | *Parafusulina* | 150 |
| Schwagerinidae | Schwagerininae | *Pseudofusulina* | 150 |
| Schwagerinidae | Schwagerininae | *Quasifusulina* | 150 |
| Schwagerinidae | Schwagerininae | *Rugosofusulina* | 150 |
| Schwagerinidae | Schwagerininae | *Schwagerina* | 150 |
| Schwagerinidae | Schwagerininae | *Triticites* | 150 |
| Schwagerinidae | Pseudoschwagerininae | *Pseudoschwagerina* | 150 |
| Ozawainellidae | Ozawainellinae | *Eostaffella* | 150 |
| Schubertellidae | Schubertellinae | *Schubertella* | 150 |
| Neoschwagerinidae | Neoschwagerininae | *Neoschwagerina* | 150 |
| Verbeekinidae | Misellininae | *Misellina* | 150 |

*Note*: Genus level is used in this work. The classification system follows that of Sheng et al. (1988).

that it can successfully predict a correct label for a given input image. The category can be species, genera or other taxonomic ranks. In this work, the goal is to predict the labels of fusulinid images to the genus level. More specifically, a set of known labelled data $\{(\mathbf{X}_{img}, y), \ldots\}$ are given for training. Each category $y \in \{1, 2, 3, \ldots, k\}$ has at least one training sample. We build model $f$ with trainable parameters and train $f$ using the available training set. After that, the trained $f$ can predict label $\hat{y}$ given an unlabelled image $\mathbf{X}_{img}$.

## 2.3 | Method

### 2.3.1 | Overview

This work introduces a multiview ensemble framework (or a meta-method) for fossil image identification. The purpose is to further improve the performance of representative deep learning models for fossil classification by using suitable computer science techniques and considering fossil characteristics. Concretely, we employ multiple base models to learn from the multiviews of the original input fossil images. Each base model and view are one-to-one correspondence so that each base model can extract diverse features for making individual predictions. The final decision is made by combining the predictions from multiple classifiers following the wisdom of crowds so that the ensemble of multiple classifiers can benefit from a more complementary set of diverse features. There could be many different choices of the multiviews of original images, and Figure 1 elaborates the proposed framework using the Original, Grey and Skeleton views (i.e. OGS method) towards fossil image classification.

### 2.3.2 | Base models

The base models are the fundamental components of the proposed framework. The qualified base models should first be the candidate for solving the problem formulated in Section 2.2. Since the problem to solve is a typical multiclass image classification problem, a large number of models based on deep convolutional neural networks (CNNs) can be adopted as the base models for this problem (Li et al., 2021). Essentially, the base model tries to automatically extract proper image features using the convolutional filters, such that these features are discriminative for making correct predictions. Each base model is trained using one view of the labelled images, and the error (via cross-entropy) between the predicted label (a predicted probability distribution over all classes) and the ground truth label (a probability distribution where 1 for the true class and 0 for other classes, i.e. one-hot vector) is back propagated to adjust the trainable parameters in deep learning models. After training, we obtain the trained base model $f_{base}$ which can map or transform an input image $\mathbf{X}_{image}$ to a probability distribution $\mathbf{z}$ overall $k$ classes, that is,

$$f_{base} : \mathbf{X}_{image} \mapsto \mathbf{z} \in \mathbb{R}^k. \quad (1)$$

The selected milestones of deep learning models for image classification are summarized below. These models, with the latest data argumentation and the state-of-the-art updates by famous deep learning library TIMM[1] and also with the pre-trained model parameters from ImageNet dataset, are respectively employed as the base model $f_{base}$ of the proposed multiview ensemble framework.

---

[1] https://github.com/huggingface/pytorch-image-models.

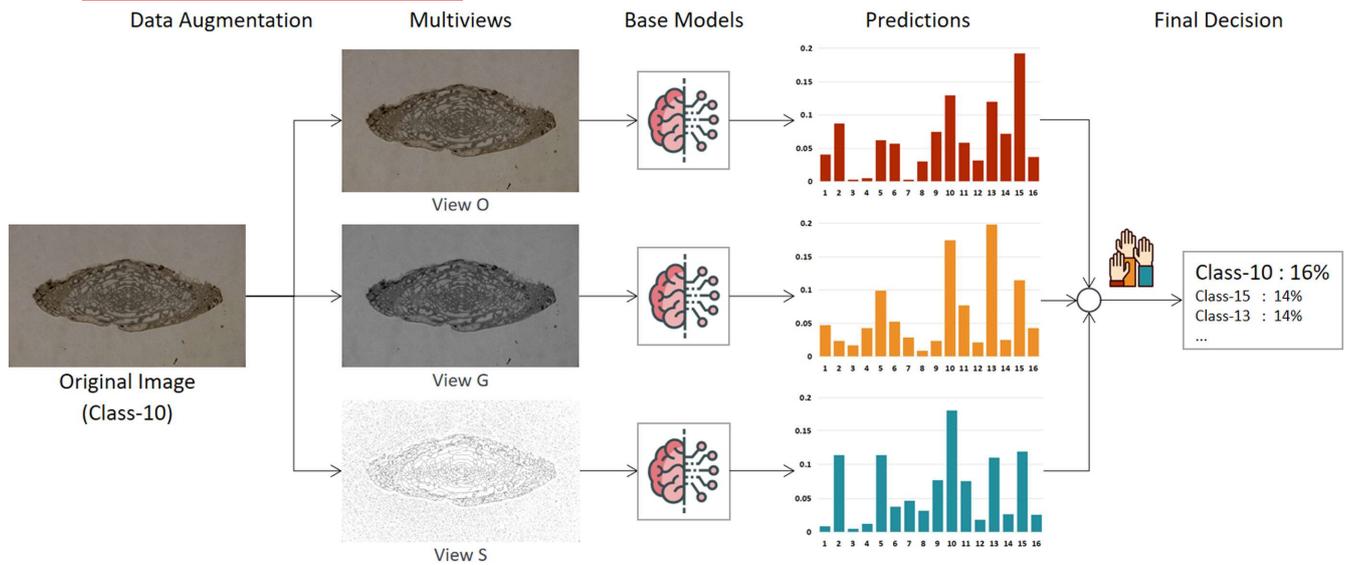

**FIGURE 1** Overview of the proposed multiview ensemble framework and the OGS method for fossil image identification. The original image is augmented into three views, that is, Original, Grey and Skeleton views. Each view is fed into respective base models. The final decision is made by combining the predictions from these base models. The example fusulinid image is *Beedeina euryteines*, courtesy of Dr. Rafael Augusto Pires de Lima.

- ResNet or Residual Network (He et al., 2016) is one of the representative types of CNNs, which aims to effectively include more convolutional layers in CNN using skip connections between some layers. The specific ResNet adopted in this work is ResNet-50.[2]
- MobileNet (Howard et al., 2017) is a light model that greatly reduces the number of parameters in CNN, and is originally designed for mobile devices. MobileNet-v3 (more specifically MobileNet-v3-large-100)[3] is tested in this work, which uses the neural architecture search (Zoph & Le, 2016) to modify MobileNet.
- Inception-v4 (Szegedy et al., 2017) is a CNN model developed from Inception-v1 also known as GoogLeNet (Szegedy et al., 2015) where the inception module is introduced. Compared to previous versions, Inception-v4[4] has a simplified architecture (without residual connections) with more inception modules.
- EfficientNet (Tan & Le, 2019) presents a novel approach to uniformly scale width, depth, and resolution over a base CNN model using a compound coefficient (given that coefficient in a constraint optimization for width, depth and resolution). Considering relatively small datasets, EfficientNet-b2[5] is chosen for the experiments.
- RegNet (Radosavovic et al., 2020) is a simple network design space coming from the neural architecture search (Zoph & Le, 2016) over a large network design space. The network design space is restricted by the quantized linear function for widths and depths. RegNetY (more specifically RegNetY-032)[6] is experimented in this work.

### 2.3.3 | Data augmented multiviews

The purpose of multiviews is to encourage base models to make good individual predictions, and in the meanwhile be complementary to each other. The classical bagging strategy (Zhou, 2021) that creates multiple random subsets of the original training set would likely reduce the available unique training data at each view. To alleviate this challenge, the most regular method is to duplicate the original training set for each view, which gives the naive version called OOO when considering three views. To increase the diversity between the three views and base models, we further propose two extra meaningful views called the Grey view (ignoring RGB colour that contains no morphological information) and the Skeleton view (focusing on the topology of fossil skeleton), and accordingly come up with the augmented version called OGS. There could be other possible meaningful transformations, other combinations of views and even many more views. We provide preliminary research in this direction and leave others as future work. Formally, we have

$$f_{trans\_1}, f_{trans\_2}, \ldots, f_{trans\_m} : \mathbf{X}_{image} \mapsto \mathbf{X}_{view\_1}, \mathbf{X}_{view\_2}, \ldots, \mathbf{X}_{view\_m}, \quad (2)$$

where the function $f_{trans\_m}$ transforms the original image $\mathbf{X}_{image}$ to view $m$ and produces augmented image $\mathbf{X}_{view\_m}$. Letting $m = 3$, for naive version OOO, $f_{trans\_1}, f_{trans\_2}, f_{trans\_3} = I, I, I$ where $I$ is identity matrix. Regarding OGS, $f_{trans\_2}$ converts original images to grey images via $Grey = 0.299R + 0.587G + 0.114B$ where $R$ is for the red channel, $G$ is for the green channel and $B$ is for the blue channel of the original images. $f_{trans\_3}$ converts the grey images to binary images, and then employs Zhang's method (Zhang & Suen, 1984) for skeletonization that reduces binary objects to one pixel-wide representation.

---

[2] https://github.com/huggingface/pytorch-image-models/blob/main/timm/models/resnet.py.
[3] https://github.com/huggingface/pytorch-image-models/blob/main/timm/models/mobilenetv3.py.
[4] https://github.com/huggingface/pytorch-image-models/blob/main/timm/models/inception_v4.py.
[5] https://github.com/huggingface/pytorch-image-models/blob/main/timm/models/efficientnet.py.
[6] https://github.com/huggingface/pytorch-image-models/blob/main/timm/models/regnet.py.

### 2.3.4 | Ensemble mechanism

To reduce the risk of overfitting due to the limited and expensive labelled fossil data, we maintain the number of unique training data at each view by data augmentation as described in Section 2.3.3, which is distinguished from bagging strategy (Dong et al., 2020). The $m$ views of original images $\mathbf{X}_{view\_m}$ are respectively fed to $m$ base models $f_{base\_m}$, and produce $m$ probability distribution $\mathbf{z}_m$ over $k$ classes. For each view and base model, we mathematically have

$$\mathbf{z}_m = f_{base\_m}(\mathbf{X}_{view\_m}), \quad \mathbf{z}_m \in \mathbb{R}^k. \tag{3}$$

Each base model is trained using the same number of training data $\{(\mathbf{X}_{view\_m}, y), \ldots\}$ from each view. The trainable parameters in $f_{base\_m}$ are optimized by gradually reducing the error between the predicted class probability distribution $\mathbf{z}_m$ and ground truth label probability distribution. After training, each base model $f_{base\_m}$ can be used to make its own predictions $\mathbf{z}_m$ given the corresponding augmented image $\mathbf{X}_{view\_m}$. The final decision is made via

$$\hat{y} = \arg\max_k \mathbf{z} = \arg\max_k f_{comb}(\mathbf{z}_1, \mathbf{z}_2, \ldots, \mathbf{z}_m), \tag{4}$$

where function $f_{comb}$ combines the predictions $\mathbf{z}_1, \mathbf{z}_2, \ldots, \mathbf{z}_m$ from $m$ views and models, and produces the final prediction or final probability distribution $\mathbf{z}$. Note that $m$ is set to three in this work, and we take soft voting, that is, $f_{comb}(\mathbf{z}_1, \mathbf{z}_2, \mathbf{z}_3) = (\mathbf{z}_1 + \mathbf{z}_2 + \mathbf{z}_3)/3$ for combining the predictions. The operator $\arg\max$ over $k$ means finding the maximum probability in vector $\mathbf{z} \in \mathbb{R}^k$ and returning the corresponding index as the predicted label $\hat{y}$.

## 2.4 | Experimental settings

The main dataset, that is, the Huang et al. (2023) dataset, consists of 2400 fossil images of fusulinid individuals of 16 genera, with 150 images each. For this multiclass classification problem, the widely used Acc@1 (true class matching with the top-1 probable predicted class, which is equivalent to Micro-F1 in our case), Acc@3 (true class included in the top-3 probable predicted classes) and Macro-F1 (harmonic mean of precision and recall over classes) are adopted as the metrics to evaluate the performance of the trained model in predicting test images.

Regarding hyperparameters, we search learning rate [0.001, 0.01, 0.1] and batch size [32, 64, 128] for each base model (in total nine combinations), and set the epoch to 500 and use the default TIMM hyperparameters for others. The best hyperparameters of each model for each view are selected respectively by comparing the average of two independent runs of Acc@1 results (see Supporting Information S2 for the best hyperparameters used). And these hyperparameters are then employed in the following experiments. The experiments are conducted on the GPU server, NVIDIA GeForce RTX 3090 Ti with 24G memory.

## 3 | RESULTS

### 3.1 | Main experiments

In the main experiments, we consider the Huang et al. (2023) dataset as described in Section 2.1, and feed the majority of data with labels to train the model. Specifically, each class has 110 images for training and 20 images each for validation and testing. Table 2 compares the OGS method to other typical variants as baselines.

Some key observations from Table 2 are as follows. First, a comparison between OOO and O shows the naive ensembles of deep learning models can further improve the performance in most cases, while OOO obtains worse results than O when taking EfficientNet-b2 as the base model. Second, compared to the O column, that is, the baseline that simply trains the base model using original images, the ensemble of data augmented multiviews OGS consistently outperforms the baseline O with gains ranging in [0.81, 2.32], [0.41, 1.28] and [0.89, 2.5] for Acc@1, Acc@3 and Macro-F1 respectively. The improvement gains of Acc@3 are less than Acc@1, since Acc@1 is a stricter metric than Acc@3 as introduced in experimental settings. Third, the ensemble of data augmented multiviews with Grey and Skeleton views, that is, OGS, can obtain generally superior performances (11 of 15 cases in terms of mean values) compared to OOO, despite the performances of OOO are already quite high. Overall, OGS generally achieves the best performance regarding all five deep learning models and three widely used multiclass classification metrics.

### 3.2 | Different ratio of train set

Machine learning models often need sufficient labelled data to train the model so as to relieve the potential overfitting issue. Nevertheless, it might be expensive or hard to annotate data, which is typical when it requires domain experts for annotation, like in our case of fusulinids. To this end, we simulate such scenarios by decreasing training data. Specifically, the ratios of images in each class for train, validation and test set are 0.1–0.8 (with step 0.1), 0.1 and 0.8–0.1 (with step 0.1) respectively, that is, eight different data splits for benchmarks. We choose ResNet-50 (OGS achieving best results in Table 2), Inception-v4 (OGS and OOO obtaining similar results in Table 2) and RegNetY (OOO achieving best results in Table 2) for the experiments, and the results are illustrated in Figure 2.

It is interesting to observe that as the train ratio decreases, OGS and OOO (under the proposed framework) obtain more performance gains compared with O that employs the original images to train a single base model, for example, OGS receives about 6% top-1 accuracy gains (on the basis of about 69%) for ResNet-50 when train ratio decreases to 0.1. Besides, the error bars of standard deviation of 10

TABLE 2 The main experimental results of O, G, S, OOO and OGS implementations over the Huang et al. (2023) dataset.

|  | O | G | S | OOO | OGS |
|---|---|---|---|---|---|
| Acc@1 |  |  |  |  |  |
| ResNet-50 | 87.98$_{\pm1.19}$ | 88.78$_{\pm0.86}$ | 83.78$_{\pm1.51}$ | 88.88$_{\pm0.80}$ | **90.30**$^\dagger{}_{\pm0.65}$ |
| MobileNet-v3 | 90.23$_{\pm1.19}$ | 90.72$_{\pm0.75}$ | 85.33$_{\pm1.01}$ | **91.33**$_{\pm0.65}$ | **91.59**$^\dagger{}_{\pm0.66}$ |
| Inception-v4 | 89.78$_{\pm1.13}$ | 89.63$_{\pm1.18}$ | 85.39$_{\pm1.52}$ | **91.16**$^\dagger{}_{\pm0.70}$ | **91.16**$_{\pm0.79}$ |
| Efficientnet-b2 | **90.72**$_{\pm0.50}$ | 90.33$_{\pm0.79}$ | 85.92$_{\pm1.47}$ | 90.48$_{\pm0.49}$ | **91.53**$^\dagger{}_{\pm0.77}$ |
| RegnetY | 90.30$_{\pm0.81}$ | 90.03$_{\pm0.92}$ | 85.23$_{\pm1.51}$ | **91.59**$^\dagger{}_{\pm0.75}$ | **91.41**$_{\pm0.81}$ |
| Acc@3 |  |  |  |  |  |
| ResNet-50 | 97.64$_{\pm0.47}$ | 97.42$_{\pm0.70}$ | 95.55$_{\pm0.73}$ | **98.08**$_{\pm0.42}$ | **98.47**$^\dagger{}_{\pm0.42}$ |
| MobileNet-v3 | 98.00$_{\pm0.58}$ | 98.02$_{\pm0.49}$ | 96.13$_{\pm0.71}$ | **98.56**$_{\pm0.30}$ | **98.64**$^\dagger{}_{\pm0.36}$ |
| Inception-v4 | 98.06$_{\pm0.58}$ | 97.98$_{\pm0.64}$ | 96.47$_{\pm0.63}$ | **98.64**$^\dagger{}_{\pm0.43}$ | **98.47**$_{\pm0.48}$ |
| Efficientnet-b2 | 97.70$_{\pm0.57}$ | 97.84$_{\pm0.71}$ | 95.72$_{\pm0.90}$ | **98.06**$_{\pm0.54}$ | **98.42**$^\dagger{}_{\pm0.44}$ |
| RegnetY | 97.03$_{\pm0.64}$ | 97.16$_{\pm0.75}$ | 96.59$_{\pm0.81}$ | **97.91**$_{\pm0.46}$ | **98.31**$^\dagger{}_{\pm0.61}$ |
| Macro-F1 |  |  |  |  |  |
| ResNet-50 | 87.73$_{\pm1.18}$ | 88.67$_{\pm0.87}$ | 83.67$_{\pm1.58}$ | 88.69$_{\pm0.81}$ | **90.23**$^\dagger{}_{\pm0.67}$ |
| MobileNet-v3 | 90.14$_{\pm1.19}$ | 90.67$_{\pm0.78}$ | 85.24$_{\pm1.03}$ | **91.25**$_{\pm0.67}$ | **91.55**$^\dagger{}_{\pm0.69}$ |
| Inception-v4 | 89.62$_{\pm1.24}$ | 89.51$_{\pm1.19}$ | 85.38$_{\pm1.55}$ | **91.03**$_{\pm0.73}$ | **91.09**$^\dagger{}_{\pm0.81}$ |
| Efficientnet-b2 | **90.55**$_{\pm0.51}$ | 90.25$_{\pm0.79}$ | 85.92$_{\pm1.47}$ | 90.31$_{\pm0.50}$ | **91.44**$^\dagger{}_{\pm0.79}$ |
| RegnetY | 90.20$_{\pm0.83}$ | 89.93$_{\pm0.94}$ | 85.28$_{\pm1.47}$ | **91.48**$^\dagger{}_{\pm0.77}$ | **91.34**$_{\pm0.79}$ |

*Note*: The abbreviations O, G and S are for the Original, Grey and Skeleton views respectively. The results along the O column are considered as the baseline that directly trains the base model using original images. The top-2 performances along each row are in bold, and the top-1 is also marked with †. Each entry describes the mean ± standard deviation obtained from 20 independent runs.

independent runs indicate that OGS and OOO are more robust than O. Furthermore, OGS and OOO reach comparable performance for various training ratios when taking RegNetY as the base model; OGS obtains much better performance than OOO and O (e.g. 1.91% and 5.24% Acc@1 gains respectively) when considering the smallest train ratio 0.1 for Inception-v4 as the base model, despite OGS and OOO obtain similar results when feeding about 0.733 of the dataset for training as shown in Table 2; OGS considerably outperforms OOO for most training ratios when taking ResNet-50 as the base model.

## 3.3 | Comparing to similar ensemble framework

The proposed multiview ensemble framework is closely related to the bagging framework. Both ensemble frameworks individually train multiple base models and make the final predictions by combining results from the multiple trained base models. They only differ in the inputs to multiple base models, but they both try to increase the diversity between base models by imposing data augmentation or sampling techniques over the inputs. Therefore, the proposed framework is compared to OOO-bagging. We follow the standard bagging framework (Zhou, 2021) to implement the OOO-bagging method. Concretely, the test set of OOO-bagging exactly follows that of OGS, that is, first randomly taking out 20 test images for each class as described in the main experiments. For OOO-bagging, the same number of the remaining images are randomly sampled with replacement from the remaining images to create the training set (hence some images might be sampled multiple times), and the final rest of the images that are not sampled act as the validation set. The OOO-bagging method repeats such bootstrapping strategy three times from the original view and accordingly trains three base models to form the ensemble. The performance comparison of the OOO-bagging and the OGS is illustrated in Figure 3.

We observe that OGS significantly outperforms OOO-bagging for ResNet-50 and Inception-v4, though they obtain comparable performance for other base models. The results indicate that the proposed framework might be more effective than the classical bagging framework given the similar computational budget of the ensemble of three base models. The potential reason could be that the number of unique training data for OGS is more than that for OOO-bagging,[7] which might therefore alleviate the overfitting issue due to insufficient training data.

---

[7]For our case, according to Sections 2.4 and 3.1, the number of unique training samples for each base model of OGS is 110 per class (or 1760 for all classes), while that of OOO-bagging is less than 110. Concretely, the bootstrapping strategy of OOO-bagging randomly takes the number of $n$ samples with replacement from the given $n$ samples. The number of unique training samples can be calculated via $1 + \frac{n-1}{n} + \left(\frac{n-1}{n}\right)^2 + \cdots + \left(\frac{n-1}{n}\right)^{n-1}$ where the first term 1 is the probability of the first sample being non-repetitive; the second term $\frac{n-1}{n}$ is the probability of the second sample being non-repetitive; the third term $\left(\frac{n-1}{n}\right)^2$ is the probability of the third sample being non-repetitive; and the last term is the probability of the $n$-th sample being non-repetitive. Applying the formula of summation for the geometric sequence, we finally derive

$$\left[1-\left(\frac{n-1}{n}\right)^n\right] \times n = \left[1-\left(\frac{130-1}{130}\right)^{130}\right] \times 130 \approx 0.6335 \times 130 = 82.355,$$

where $n = 130$ since there are 20 images per class reserved as test set, the number of unique training samples is around 82 images per class (or around 1318 images for all classes), and the remaining 48 images per class are used as the validation set. Therefore, the number of unique training samples to each base model for OOO-bagging is around 1318 for all classes, which is smaller than OGS with 1760 unique training samples, that is, reducing roughly 25% non-repetitive training samples.

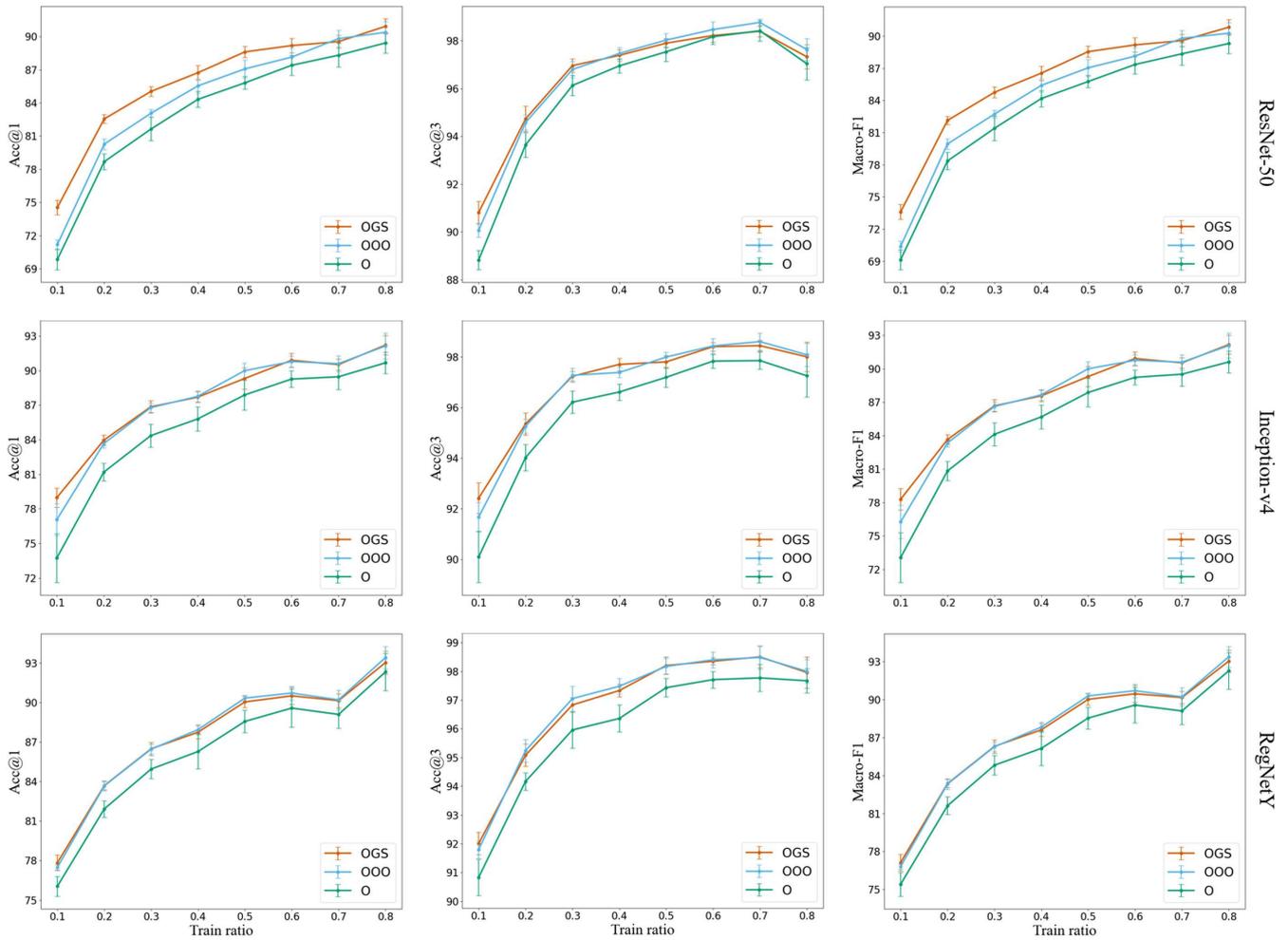

**FIGURE 2** The results of Acc@1, Acc@3 and Macro-F1 (from left to right) under various train ratios (0.1–0.8 with step 0.1) for the base model ResNet-50 (row 1), Inception-v4 (row 2) and RegNetY (row 3). Best viewed in colours.

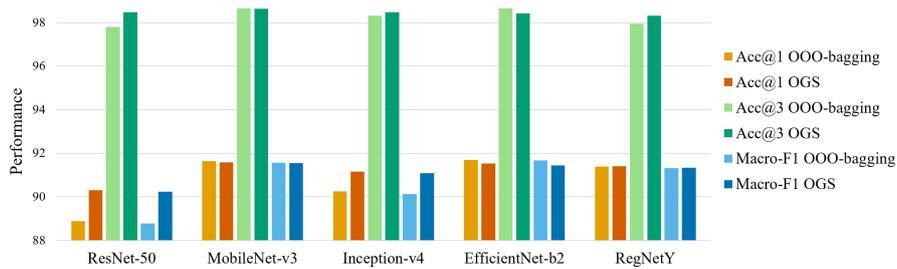

**FIGURE 3** The proposed framework versus bagging framework. Regarding the stricter metrics of Acc@1 and Macro-F1, OGS under the proposed framework significantly outperforms OOO-bagging under the classical bagging framework for ResNet-50 and Inception-v4. However, they obtain comparable performance for Acc@3 and for other base models. Best viewed in colours.

## 4 | DISCUSSION

### 4.1 | The proposed three views and OGS method

According to Table 2, OGS generally achieves the best performances compared to other methods regarding all the five types of base models and all three metrics, and the ensemble OGS consistently outperforms the baseline O that trains a single base model. Table 2 also suggests the superior or comparable performance of OGS against OOO, that is, the data augmented three views would gain more benefits compared to the duplicated three views. The reason could owe to the improvement of the diversity of the predictions among three base models when the data augmented three views are fed to the three base models.

To support this claim, we plot the confusion matrices of the Original view, the Grey view, the Skeleton view and the multiview ensemble for ResNet-50 as an example, shown in Figure 4. It can be observed that different prediction patterns are generated by different views, with the ensemble model obtaining the optimal performance in terms of accuracy (i.e. the sum of the numbers along the diagonal being the greatest). For example, the correct predictions of Triticites for respective views (O, G and S) are 17, 16 and 13, and the proposed framework boosts the performance to 19. This points out that the different views bring in inference diversity, and the ensemble procedure is able to revise misidentification by the single model.

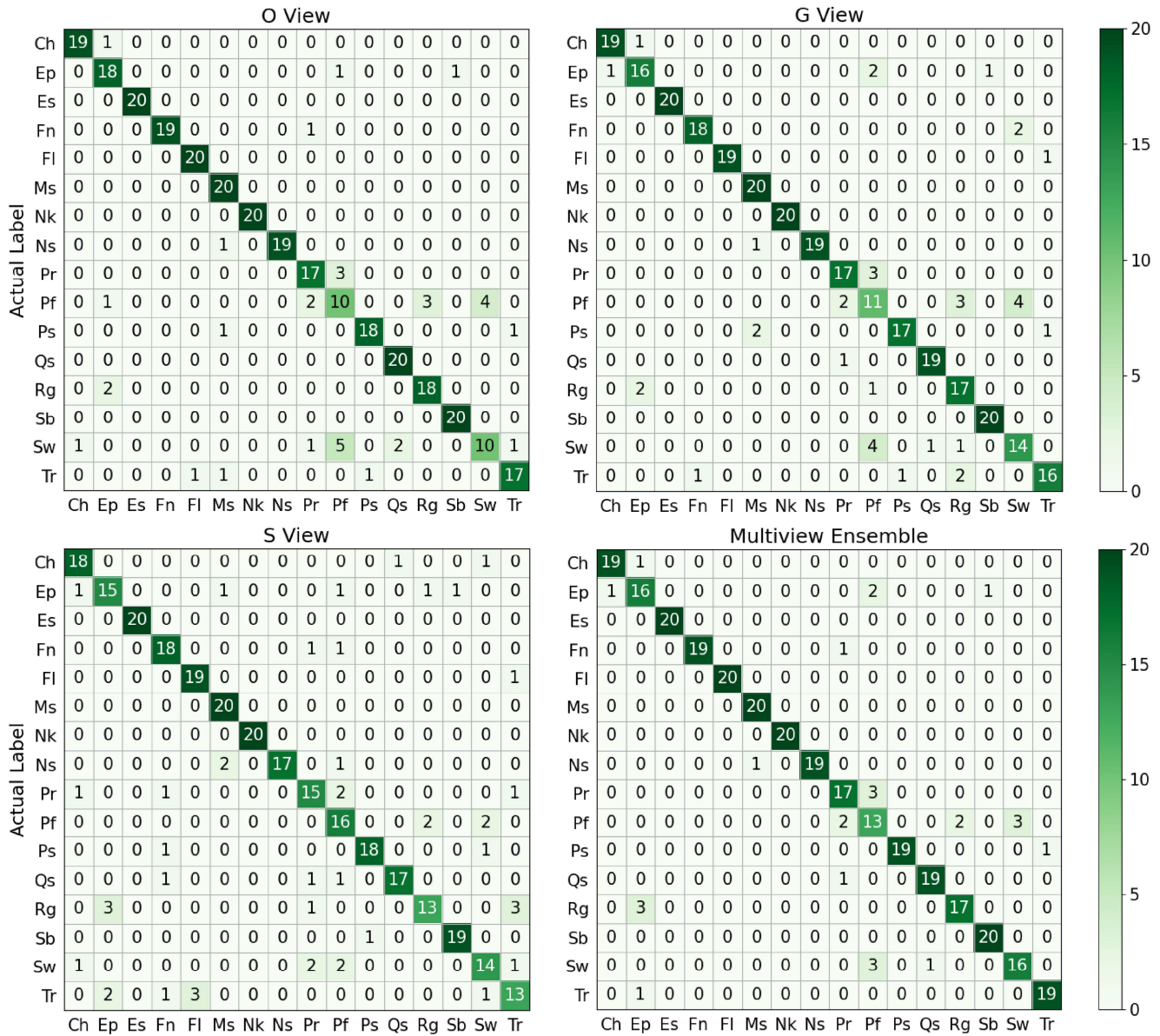

**FIGURE 4** The confusion matrix for the Original view, the Grey view, the Skeleton view and the multiview ensembles when the base model is ResNet-50. The deeper colour indicates the larger number; the maximum number is 20 (the number of test set images of each genus). The number along the diagonal line represents the number of corrected predictions, that is, the predicted label (*x*-axis) matches the actual label (*y*-axis). Best viewed in colours. Ch, *Chusenella*; Ep, *Eoparafusulina*; Es, *Eostaffella*; Fn, *Fusulina*; Fl, *Fusulinella*; Ms, *Misellina*; Nk, *Nankinella*; Ns, *Neoschwagerina*; Pr, *Parafusulina*; Pf, *Pseudofusulina*; Ps, *Pseudoschwagerina*; Qs, *Quasifusulina*; Rg, *Rugosofusulina*; Sb, *Schubertella*; Sw, *Schwagerina*; Tr, *Triticites*.

The diversity in views leads to their diversity in the extraction of fossil structural information. This is especially clear when it comes to the subfamily Schwagerininae (see the taxonomy in Table 1) and can be confirmed by class activation mapping (CAM). This method weighs the sum of the presence of visual patterns at different spatial locations and underlines the image regions most relevant to a particular category (Zhou et al., 2015).

Figure 5 shows the visualization of Grad-CAM (Selvaraju et al., 2016) of O, G, and S on images of five individuals from the subfamily Schwagerininae, using timm-vis.[8] The first two views, O and G, tend to activate in the middle of the image, which is the region of the proloculus (the very first coiling whorl of fusulinids) and the two to three inner whorls. The S view, on the other hand, is often dispersed and activates at the periphery of the fusulinid, which is the whorls grown at the later stage. Figure 4 shows that the S view particularly achieves better performance on the two schwagerine groups *Pseudofusulina* and *Schwagerina*, compared to the other two views (see the 10th and 15th diagonal elements of confusion matrices). These two fusulinid genera are morphologically very similar but distinguishable for the ontogenetic development process. The proloculus of *Pseudofusulina* is oftentimes large and the succeeding whorls coil very loosely, while the proloculus of *Schwagerina* is often small with the two to three succeeding whorls coiling intensely and gradually loosening outwards. Although the differences are distinct with regard to their development and fusulinid taxonomists agree to put them into two genera (Moore, 1964; Sheng et al., 1988), controversies exist in the identification on a case-by-case basis (see part 4.3 for a consistency estimation). The skeleton view, as seen in the CAM result, may be able to highlight the differences in proloculus and outer whorls and therefore shows more excellence in these subtly distinguishable groups. In the case of ensemble learning, multiple views can complement each other by highlighting different features. Although the Grad-CAM results may not fully reflect the model's 'attention' distribution, they are a good demonstration of the fact that a simple manipulation of fossil images (such as the skeletonization) can emphasize unique features of the identical individual for the model, so that the ensemble model can synthesize the classification information obtained from more aspects and reach the better results.

---

[8] https://github.com/novice03/timm-vis.

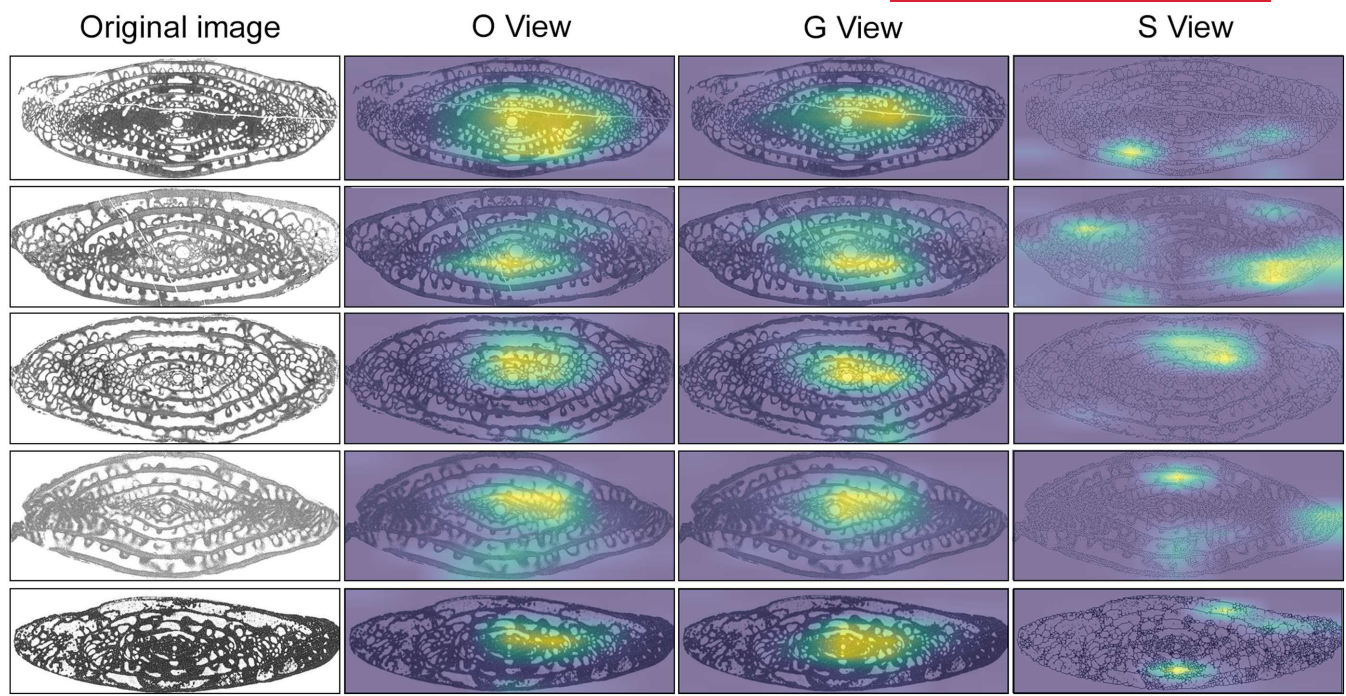

**FIGURE 5** Examples of original images and corresponding visualization results of activation mapping of the Original view, the Grey view and the Skeleton view (left–right), generated by Grad-CAM (Selvaraju et al., 2016). The base model is ResNet-50, consistent with the model used in Figure 4. The five input images are from the species of subfamily Schwagerininae including *Parafusulina australis*, *Rugosofusulina mansuyi*, *Pseudofusulina wulungensis*, *Schwagerina neoaculata* and *Triticites kawensis* (top-down). These images are not to scale. The contribution of different regions to identification is indicated by a colour ranging from blue to yellow. The yellow highlighted regions contribute the most. Note that the activated regions of the S view are clearly distinct from those of the other two, indicating different features detected and analysed. Best viewed in colours.

It might be worth mentioning that this work proposes to utilize O, G and S views to form the ensemble, but there could be other possible views that researchers or engineers can further explore based on the characteristics of their fossil images. It is also the case that the proposed framework or meta-method can be directly used or easily modified for broader fossil or extant organisms image classification problems.

### 4.2 | High applicability to small datasets

In Figure 2, the most significant finding is that as the training ratio of labelled images decreases, the proposed framework, especially the OGS method, generally receives more performance gains compared to simply using one single model. This finding indicates that the proposed framework has a substantial application when the labelled data are insufficient. Fossil image data often fit into this category, which is oftentimes limited as a result of fossil preservation, sampling intensity and a requirement of domain-specific knowledge for fossil image annotation. Consequently, lacking labelled training data is a common challenge in fossil image identification and hinders the advancement of automatic identification methods. The proposed multiview ensemble framework is proven to be likely to perform much better than simply applying a single model provided insufficient training data, thus showing the promise for similar practices on other fossils, especially rare ones.

TABLE 3 The experimental results of O, G, S, OOO and OGS implementations over the Pires de Lima et al. (2020) dataset, which has the class imbalanced characteristic and is smaller than Huang et al. (2023) dataset in Table 2.

|  | O | G | S | OOO | OGS |
|---|---|---|---|---|---|
| ACC@1 |  |  |  |  |  |
| ResNet-50 | $81.57_{\pm4.67}$ | $82.07_{\pm3.48}$ | $70.64_{\pm3.66}$ | $85.21^{\dagger}_{\pm3.57}$ | $85.14_{\pm2.8}$ |
| MobileNet-v3 | $83.57_{\pm3.57}$ | $83.29_{\pm4.74}$ | $73.57_{\pm3.46}$ | $87.07_{\pm3.45}$ | $87.50^{\dagger}_{\pm2.51}$ |
| Inception-v4 | $86.29_{\pm2.57}$ | $86.21_{\pm4.32}$ | $75.57_{\pm3.78}$ | $85.79_{\pm2.23}$ | $88.79^{\dagger}_{\pm3.04}$ |
| Efficientnet-b2 | $86.29_{\pm3.11}$ | $82.57_{\pm3.80}$ | $69.50_{\pm4.34}$ | $87.64^{\dagger}_{\pm1.65}$ | $87.36_{\pm3.04}$ |
| RegnetY | $84.57_{\pm2.29}$ | $85.14_{\pm3.36}$ | $73.21_{\pm4.33}$ | $85.29_{\pm1.57}$ | $89.07^{\dagger}_{\pm2.57}$ |
| ACC@3 |  |  |  |  |  |
| ResNet-50 | $96.21_{\pm1.77}$ | $96.29_{\pm1.99}$ | $93.71_{\pm2.53}$ | $97.71_{\pm1.46}$ | $97.86^{\dagger}_{\pm1.53}$ |
| MobileNet-v3 | $97.14_{\pm1.56}$ | $95.64_{\pm2.14}$ | $93.00_{\pm2.16}$ | $98.07^{\dagger}_{\pm1.22}$ | $97.43_{\pm1.25}$ |
| Inception-v4 | $97.71_{\pm1.46}$ | $98.71_{\pm1.10}$ | $94.71_{\pm2.39}$ | $98.29_{\pm0.86}$ | $99.93^{\dagger}_{\pm0.31}$ |
| Efficientnet-b2 | $97.79_{\pm1.60}$ | $97.93_{\pm1.78}$ | $90.71_{\pm3.27}$ | $98.14_{\pm1.12}$ | $98.50^{\dagger}_{\pm1.46}$ |
| RegnetY | $96.93_{\pm1.58}$ | $97.36_{\pm1.71}$ | $93.57_{\pm2.24}$ | $98.14_{\pm1.2}$ | $98.50^{\dagger}_{\pm1.24}$ |
| Macro-F1 |  |  |  |  |  |
| ResNet-50 | $82.24_{\pm4.06}$ | $83.19_{\pm3.71}$ | $72.08_{\pm4.25}$ | $86.09^{\dagger}_{\pm2.83}$ | $85.78_{\pm2.88}$ |
| MobileNet-v3 | $84.05_{\pm3.29}$ | $84.48_{\pm4.91}$ | $75.52_{\pm3.05}$ | $87.17_{\pm3.21}$ | $88.58^{\dagger}_{\pm2.54}$ |
| Inception-v4 | $86.62_{\pm1.90}$ | $87.08_{\pm4.04}$ | $78.08_{\pm3.37}$ | $86.21_{\pm1.67}$ | $89.05^{\dagger}_{\pm2.93}$ |
| Efficientnet-b2 | $86.53_{\pm3.44}$ | $83.80_{\pm3.68}$ | $70.46_{\pm4.51}$ | $87.85_{\pm1.56}$ | $88.33^{\dagger}_{\pm3.00}$ |
| RegnetY | $84.90_{\pm2.14}$ | $86.54_{\pm3.09}$ | $75.21_{\pm4.37}$ | $85.32_{\pm1.52}$ | $89.83^{\dagger}_{\pm2.10}$ |

To further demonstrate the applicability to small datasets, we follow the experimental settings as used in the main experiments[9] to conduct experiments on a much smaller real-world fossil dataset, that is, the Pires de Lima et al. (2020) dataset. This dataset contains 342 sectioned fusulinid images assigned to eight genera and exhibits a significant category imbalance. There are 88 images in the largest genera and only 15 in the smallest. Unlike the Huang et al. (2023) dataset, images in this dataset are not segmented, and thus background is preserved. Also, the colour difference across different image sources is more prominent. The experimental results are shown in Table 3.

Comparing to the main experiments over the larger dataset (Huang et al., 2023), more considerable improvements of OGS against OOO are presented for the experiments over this smaller dataset (Pires de Lima et al., 2020) when taking Inception-V4 (+3% for ACC@1 and +2.84% for Macro-F1) and RegnetY (+3.78% for ACC@1 and +4.51% for Macro-F1) as the base in Table 3, though OGS has shown considerable improvements when taking ResNet-50 (+1.42% for ACC@1 and +1.52% Macro-F1) and Efficientnet-b2 (+1.05% for ACC@1 and +1.13% Macro-F1) as the base over the larger dataset in Table 2. The reason could be that the larger dataset provides relatively sufficient training data and that the performances are already saturated, while the smaller and imbalance dataset (Pires de Lima et al., 2020) offers a more challenging classification scenario to reveal the superiority of OGS against OOO better. On the other hand, it might not be a fair comparison between this work and Pires de Lima et al. (2020) work over the same dataset due to the lack of access to their model hyperparameters and detailed training schemes. Nonetheless, for top-1 accuracy of ResNet-50, the only model used in both studies, OGS (85.14%) obtains a significant improvement compared with O (81.57%) as well as the model of their work (80%).

The mechanism by which the O, G and S views help the ensemble model obtain the correct identification result is worth exploring. Fossils are preserved in sedimentary rocks, and the chemical composition of fossils can be greatly affected by the surrounding rocks and fluids during taphonomic processes (Behrensmeyer et al., 2000; Martin, 1999), and differences in the composition can produce different colours that do not contain information regarding the fossil structure itself. Although the colour may somewhat reflect differences in the living and preservation environments of various classes of fossils, the optical microscope used, the filming equipment and parameters, and the factors of printing and scanning may also introduce colour-related noise. For the main dataset Huang et al. (2023), there is little difference in whether the colour is included or not, as the performance of O and G does not show much comparable variance (see Table 2). However, for the Pires de Lima et al. (2020) dataset as shown in Table 3, the Grey view performs better than the Original view for 10 of 15 cases, indicating a performance gain when colour noise is erased. As is mentioned, the Pires de Lima et al. (2020) dataset contains images of different colour schemes (may correspond to different image sources), while the Huang et al. (2023) dataset is more colour pattern consistent. Removing colour may result in greater differences (good or bad) for other potential fossil groups and thus requires caution. Conversion to skeletonized images is also useful as it helps represent the morphological structure of the fossils and thus serves as a feature extraction (Saha et al., 2016; Weeks et al., 2023). For shell-forming organisms like fusulinids, the topology of their shell, such as the number and size of chambers and the manner of spinning and coiling, is sufficient to provide a great deal of information for their identification and classification (Ross & Ross, 1991; Sheng et al., 1988; Vachard et al., 2010). Skeletonization can be seen as a feature extractor based on this prior knowledge to help the model learn the morphological features of the fossil.

---

[9]The experimental settings for the main experiments on Huang et al. (2023) dataset, and this experiments for Pires de Lima et al. (2020) dataset are the same as presented in Section 2.4, except for the blocksize hyper-parameter when performing skeletonization. The blocksize 41 is used for the Huang et al. (2023) dataset, but 61 is used for the Pires de Lima et al. (2020) dataset.

## 4.3 | Identification consistency estimation

Another point to note is the label inconsistency, which may also be responsible for the misidentification of the model. This study considers supervised classification, that is, each image needs to be labelled before training, so whether the labels can consistently indicate the features critical for classification will greatly affect the model performance. As mentioned earlier, the identification or classification of fossil species requires corresponding domain knowledge, and different experts use different morphological criteria due to different experience, training, and access to samples (Fenton et al., 2018; MacLeod et al., 2007, 2010), which leads to inconsistency in the labels they give. This inconsistency can be partially resolved by recalibration by an individual expert. Still, even self-consistency (consistency in results obtained from multiple practices of identification on the same sample by the same person) of the experts is not necessarily high (Culverhouse et al., 2014; Fenton et al., 2018). In the present study, the main dataset used contains images from multiple sources, which may introduce label inconsistency despite the fact that the dataset has been subjected to some quality control (e.g. preferential use of holo- and paratype specimens). A consistency test is performed to explore the nature of such consistency within the dataset.

Among the current 16 genera in the Huang et al. (2023) dataset, the eight genera from the Family Schwagerinidae are the most controversial groups. Inconsistency among certain genera, such as *Pseudofusulina* versus *Schwagerina* (Shamov, 1958; Shamov & Shcherbovich, 1949), *Schwagerina* versus *Chusenella* (Stewart, 1963) and *Pseudofusulina* versus *Triticites* (Shi et al., 2008), largely exist in identifications. Therefore, 160 images of these eight genera were selected for the consistency test, with two human experts involved. The original identification (which can be seen as the collective ideas of many experts, denoted as O-Label), the inference output of OGS and the two human expert re-identification results are compared. The consistency rate between two identification results of *n* images is defined as $n_{con} / n$, where $n_{con}$ is the number of images for which they present consistent labels. Table 4 shows the consistency matrix, where the consistency between two experts is the least at merely 53%, while OGS reaches the utmost consistency compared to all other inferences (85%, 68% and 58%, with O-Label and two experts respectively). This suggests that despite the many contradictions in specimen identifications among experts, the ensemble model still successfully captures the common features indicated by their collective ideas to a high degree. This shows the potential of using deep learning models to bridge contradictory and resolve inconsistency.

If a model has undergone multiple thorough training attempts, its performance should provide a quantitative assessment of the dataset's consistency. Table 5 summarizes such the assessment, where three OGS models are trained on the same images, but are provided with different labels from the original dataset and the two experts. The training and validation process follows fivefold cross-validation, that is, all samples are randomly divided into five subsets, and the ensemble models are trained over four subsets and validated over the remaining one each time until every subset has been used as the validation set once. The consistency in Table 5 is indicated using the top-1 accuracy (the mean over 10 repeated runs for each fold and then over the fivefold) between the predicted labels (by each of the three trained OGS model) and the 'ground truth' labels (when treating the labels from original, expert 1 and expert 2 as the ground truth respectively); thus rendering nine results. Surprisingly, the original labels (representing the collective ideas from multiple experts) reach the greatest self-consistency of around 67.88%, surpassing those of the two experts, though the confidence intervals may overlap. This may partly be due to the fact that the original labels are given by experts who have access to the samples, and more information like sizes and detailed structures can be acquired by close examination. Nevertheless, the relatively low consistency of these schwagerine genera points out that the taxonomy and/or classification systems are in need of reconsideration and unification.

**TABLE 4** The consistency rates between original labels (O-Label), identification results obtained by the OGS model and two human experts.

|           | O-Label | OGS  | Expert 1 | Expert 2 |
|-----------|---------|------|----------|----------|
| O-Label   | 1       | 0.85 | 0.68     | 0.57     |
| OGS       |         | 1    | 0.68     | 0.58     |
| Expert 1  |         |      | 1        | 0.53     |
| Expert 2  |         |      |          | 1        |

*Note*: The consistency test is performed on the set of eight genera of the family Schwagerinidae (*Eoparafusulina*, *Parafusulina*, *Pseudofusulina*, *Pseudoschwagerina*, *Quasifusulina*, *Rugosofusulina*, *Schwagerina* and *Triticites*), with 20 images each. The OGS model is aligned with that of Figures 4 and 5.

**TABLE 5** The consistency of the labels given in the original dataset (O-Label) and the two experts, indicated by the top-1 accuracy of OGS ensemble models on validation sets.

| Model trained on \ Ground Truth | O-Label | Expert 1 | Expert 2 |
|---|---|---|---|
| O-Label  | **67.88**$_{\pm 4.95}$ | 58.19$_{\pm 7.23}$ | 52.27$_{\pm 4.12}$ |
| Expert 1 | 63.14$_{\pm 5.15}$ | **64.22**$_{\pm 3.69}$ | 49.3$_{\pm 4.25}$ |
| Expert 2 | 49.73$_{\pm 6.86}$ | 45.24$_{\pm 5.9}$ | **59.47**$_{\pm 4.53}$ |

*Note*: The OGS models are trained using the fivefold cross-validation process to gain mean accuracy and confidence intervals.

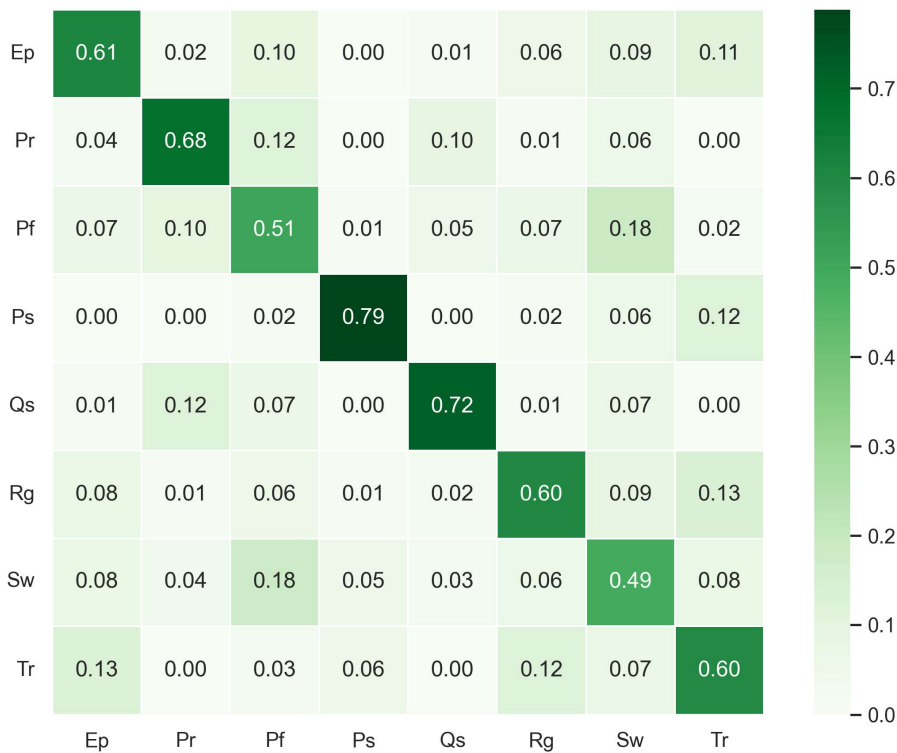

**FIGURE 6** The agreement matrix of each genus obtained in the consistency test of Table 5. Values are the expectations of labels (columns) being assigned to a specimen by an expert, given a prior assignment of labels (rows). The diagonal represents the consistency of each genus. Ep, *Eoparafusulina*; Pr, *Parafusulina*; Pf, *Pseudofusulina*; Ps, *Pseudoschwagerina*; Qs, *Quasifusulina*; Rg, *Rugosofusulina*; Sw, *Schwagerina*; Tr, *Triticites*.

On the other hand, we can look into the problem from the perspective of different genera. Based on an identification result, assuming the sampling of the specimens and the experts are random, a label 'agreement' expectation, that is, the expectation that given a label, an expert agrees to classify the specimen into the same or different label, can be calculated (see Supporting Information S1 for the computing method). Figure 6 shows the agreement matrix based on the best-performing OGS models trained on three sets of labels, including the original, expert 1 and 2 re-identified labels. The OGS models can be seen as the classification systems reflected by the three sets of labels, in which some inherent contradictions are bridged in the training process. Echoed by the actual expert identification practices, *Pseudofusulina* and *Schwagerina* hold the worst consistency of around 50%, and they are often identified as one another (18%). Other confusions exist between *Eoparafusulina* and *Triticites*, as well as *Rugosofusulina* and *Triticites*, both around 12%. Genera other than *Pseudofusulina* and *Schwagerina* have relatively better consistencies of around or over 60%. The consistency of *Pseudoschwagerina* is the largest (79%), probably because it has distinct characters like spherical test shape and tight inner whorls (Sheng et al., 1988), and thus suffers less from taxonomic controversies.

This current routine provides an approach to assess the consistency of the labels quantitatively, both across experts and categories. The results urge that the consistency of fossil identifications should be emphasized, and analysis and revision of labels should be considered before feeding data for model training. For future work, the inconsistency of fusulinids and other fossils can be better estimated and resolved by various methods, including expert-guided feature extraction of neural networks.

## 5 | CONCLUSIONS

Fossil identification is essential for evolutionary studies. Automatic identification models, especially recent advances based on deep learning, rely heavily on the quantity and quality of labelled images to train the models. However, the images are particularly limited for palaeontologists due to the fossil preservation, conditioned sampling and expensive and inconsistent label annotation by domain experts. To address these challenges, we proposed a multiview ensemble framework that collects the multiple views of each fossil specimen image reflecting its different characteristics to train multiple base models and then makes the final decision via soft voting. Regarding the characteristics of fossil images, we further proposed the Original, the Grey and the Skeleton views to establish the OGS method for identifying fossil images and conducted a case study on the fusulinid datasets.

The extensive experiments on the Huang et al. (2023) dataset as well as the Pires de Lima et al. (2020) dataset demonstrated the superiority of the proposed framework and OGS method from various aspects. In future work, it is worth investigating adopting heterogeneous base models for the proposed framework, employing other techniques to combine the outputs of base models, and experimenting on more fossil datasets. Furthermore, the consistency test showed that the proposed method could successfully integrate the ideas of multiple experts and reach the greatest consistency. The proposed routine using the performance of OGS models trained on labels provided by different experts provides an approach to assess the consistency of the labels quantitatively, both across experts and categories. These additional experiments suggest the potential application of the proposed method for assessing and resolving the inconsistencies in fossil identification.

## AUTHOR CONTRIBUTIONS

Chengbin Hou and Xinyu Lin conceived the ideas, designed the method, wrote the code and conducted the experiments; Hanhui Huang and Yukun Shi collected the data; Xinyu Lin, Hanhui Huang, Chengbin Hou and Yukun Shi analysed the data; Hairong Lv offered the computational resources; Junxuan Fa, Yukun Shi and Hairong Lv initiated the research idea; Chengbin Hou, Hanhui Huang and Xinyu Lin wrote the manuscript; Yukun Shi, Hairong Lv, Sheng Xu and Junxuan Fan provided supervision and valuable comments, and revised the manuscript. All authors contributed critically to the drafts and gave final approval for publication.

## ACKNOWLEDGEMENTS

We thank Dr. Rafael Augusto Pires De Lima for providing a fusulinid image dataset for our test, and the anonymous reviewers for their valuable advice. This work was funded by the National Natural Science Foundation of China under the Grant 42050101 and 42250104, the National Key R&D Program of China under the Grant 2021YFB3600401 and the Fujian Provincial Natural Science Foundation under the Grant 2021J01586. This is also a contribution to the IUGS Deep-time Digital Earth (DDE) Big Science Program.

## CONFLICT OF INTEREST STATEMENT

The authors declare no competing interests.

## DATA AVAILABILITY STATEMENT

The main dataset of Huang et al. (2023) used in this study can be downloaded from the DDE Repository at https://doi.org/10.12297/dpr.dde.202211.5. The dataset of Pires de Lima et al. (2020) can be obtained by requesting the original author. The source code is available at https://github.com/houchengbin/Fossil-Image-Identification, and the version producing the results in this study is archived at https://doi.org/10.5281/zenodo.8358575.

## ORCID

*Chengbin Hou*  https://orcid.org/0000-0001-6648-793X
*Xinyu Lin*  https://orcid.org/0000-0001-5164-9457
*Hanhui Huang*  https://orcid.org/0000-0002-3743-4844
*Sheng Xu*  https://orcid.org/0000-0001-6691-0857
*Junxuan Fan*  https://orcid.org/0000-0001-9913-0865
*Yukun Shi*  https://orcid.org/0000-0002-1412-179X
*Hairong Lv*  https://orcid.org/0000-0003-1568-6861

## SUPPORTING INFORMATION

Additional supporting information can be found online in the Supporting Information section at the end of this article.

**Supporting Information S1:** Calculation of agreement matrix.

**Supporting Information S2:** The best hyperparameters in the main experiments.

---

**How to cite this article:** Hou, C., Lin, X., Huang, H., Xu, S., Fan, J., Shi, Y., & Lv, H. (2023). Fossil image identification using deep learning ensembles of data augmented multiviews. *Methods in Ecology and Evolution*, *14*, 3020–3034. https://doi.org/10.1111/2041-210X.14229